\documentclass[sn-mathphys,Numbered]{sn-jnl}

\usepackage{graphicx}%
\usepackage{multirow}%
\usepackage{amsmath,amssymb,amsfonts}%
\usepackage{amsthm}%
\usepackage{mathrsfs}%
\usepackage[title]{appendix}%
\usepackage{xcolor}%
\usepackage{textcomp}%
\usepackage{manyfoot}%
\usepackage{booktabs}%
\usepackage{algorithm}%
\usepackage{algorithmicx}%
\usepackage{algpseudocode}%
\usepackage{listings}%

\usepackage{xhfill}
\usepackage{array}
\usepackage{varwidth} 
\usepackage{booktabs}
\usepackage[utf8]{inputenc}
\usepackage{amsmath}
\usepackage{booktabs} 

\usepackage{setspace}
\usepackage{lineno}

\begin{document}

\title[Article Title]{Multi-task Transfer Learning for the Prediction of Entity Modifiers in Clinical Text: Application to Opioid Use Disorder Case Detection \\
}

\author*[1,2]{\fnm{Abdullateef I.} \sur{Almudaifer}}\email{lateef11@uab.edu}
\author[1]{\fnm{Whitney} \sur{Covington}}
\author[1]{\fnm{JaMor} \sur{Hairston}}
\author[1]{\fnm{Zachary} \sur{Deitch}}
\author[1]{\fnm{Ankit} \sur{Anand}} 
\author[1]{\fnm{Caleb M.} \sur{Carroll}}
\author[1]{\fnm{Estera} \sur{Crisan}}
\author[1]{\fnm{William} \sur{Bradford}}
\author[1]{\fnm{Lauren} \sur{Walter}}
\author[1]{\fnm{Eaton} \sur{Ellen}}
\author[1]{\fnm{Sue S.} \sur{Feldman}}
\author*[1]{\fnm{John D.} \sur{Osborne}\email{ozborn@uab.edu}}

%


\affil[1]{\orgname{University of Alabama at Birmingham}, \orgaddress{\street{1720 University Blvd}, \city{Birmingham}, \postcode{35294}, \state{Alabama}, \country{United States of America}}}

\affil[2]{\orgdiv{College of Computer Science and Engineering}, \orgname{Taibah University}, \orgaddress{\city{Yanbu}, \state{Madinah}, \country{Kingdom of Saudi Arabia}}}



\abstract{
\textbf{Background:} The semantics of entities extracted from a clinical text can be dramatically altered by modifiers, including entity negation, uncertainty, conditionality, severity, and subject. Existing models for determining modifiers of clinical entities involve regular expression or features weights that are trained independently for each modifier.

\textbf{Methods:} We develop and evaluate a multi-task transformer architecture design where modifiers are learned and predicted jointly using the publicly available SemEval 2015 Task 14 corpus and a new Opioid Use Disorder (OUD) data set that contains modifiers shared with SemEval as well as novel modifiers specific for OUD. We evaluate the effectiveness of our multi-task learning approach versus previously published systems and assess the feasibility of transfer learning for clinical entity modifiers when only a portion of clinical modifiers are shared.

\textbf{Results:} Our approach achieved state-of-the-art results on the ShARe corpus from SemEval 2015 Task 14, showing an increase of 1.1\% on weighted accuracy, 1.7\% on unweighted accuracy, and 10\% on micro F1 scores.

\textbf{Conclusions:} We show that learned weights from our shared model can be effectively transferred to a new partially matched data set, validating the use of transfer learning for clinical text modifiers.
}




\maketitle

\section{Background}\label{sec1}
Modifier prediction can be considered a subset of relation extraction where a limited set of modifying relations or attributes are predicted for a fixed entity type. Modifier prediction is a critical task in information extraction since the context of a particular mention can radically alter its meaning. In clinical text, the majority of frequently used clinical observations are "negated" \citep{chapman2001evaluation}, a problem which has led to the development of algorithms such as NegEx \citep{chapman2001simple}. This problem is not unique to English and additional implementations in languages besides English have been developed \citep{chapman2013extending,mirzapour2021french}. Besides negation, algorithms have been developed to determine historical and hypothetical modifiers of clinical entities, as well as determine if the modified entity is referring to the patient or some other person. A well known example is the ConText algorithm \citep{chapman2007context}, for which there are multiple implementations \citep{chapman2007context,harkema2009context,shi2018trie,mirzapour2021french}. 

Within clinical text, algorithms for identifying modifier types associated with a particular class of clinical entities have also been developed, such as extraction of disease or medication modifiers. These algorithms can identify modifiers separately from the recognition of the clinical named entities or jointly, often with separate tracks at shared tasks to evaluate both approaches\citep{jagannatha2019overview,elhadad-etal-2015-semeval}. 
Modifiers for medication mentions were included as part of the MADE 1.0 shared task and data set\citep{jagannatha2019overview} and included dosage, route, duration, and frequency. Disease or problem specific modifiers have included modifiers for severity, a generic text, conditionality, and anatomical location. All of these modifiers were included in the SemEval 2015 Task 14 data set \citep{elhadad-etal-2015-semeval}, and subsets of them can be extracted by tools such as MedLEE \citep{friedman1999natural} (negation, uncertainty, and severity), ConText implementations \citep{chapman2007context,harkema2009context,shi2018trie,mirzapour2021french} and cTAKES (negation, body site, severity) \citep{Savova:2010hy}. What is not known is how effective transfer learning is on the detection of these modifiers. Given the wide range of modifiers that could be applied to an entity such as disease, an ideal transfer learning solution would be able to leverage previous data sets to identify existing and novel disease modifiers in a new data set.

\subsection{Modifier Classification Problem}
A more formal definition is as follows. Let $X$ be a string sequence consisting of $n$ tokens  $x_1, x_2, \dots, x_n$. Let $E=\{e_1, e_2, \dots, e_s\}$ be a string of entity mentions within $X$. Let ${m_i}  \in {M}$ be a set of predefined modifiers types of modifier $m_i$. The input to the problem is a combination of the string sequences ($E$ and the context around it $X$), denoted as $\widehat{X}$. The modifier classification problem would be, for every input sequence $\widehat{X}$, predict a modifier type $y_{m_i}(\widehat{X}) \in {m_i}$ for every modifier ${m_i}  \in {M}$ for any and all that exists.

\subsection{Related Work} \label{Related_work}
Like other areas of clinical natural language processing (NLP), work on identifying modifiers of clinical entities has been hindered by the lack of readily available public data sets. Despite this, researchers have studied a variety of methods for solving the problem of entity modifier identification, primarily using the publicly available SemEval 2015 Task 14 data set \citep{elhadad-etal-2015-semeval}, released as part of the ShARe corpus which contains modifiers for clinical problems.
As discussed in the Background section, early work on modifier identification focused only on utilizing rule-based systems \citep{Savova:2010hy, friedman1999natural, harkema2009context,shi2018trie}. Work in this area has continued, improving the speed and efficiency of modifier identification through systems such as FastContext \citep{shi2018trie} and applying this to clinical text in other languages such as French\citep{mirzapour2021french}. This system had an average an average F-Score of 0.86 F1 on electronic clinical records for negation, temporality, and experiencer. However, the trend has been to move away from an exclusively rule-based approach. For example, an early version of the open source Clinical Text Analysis and Knowledge Extraction System (cTAKES) \citep{Savova:2010hy} used exclusively a rule-based approach to predict clinical entity modifiers, but this capability of cTAKES was expanded when Dligach et al. \citep{dligach2014discovering} implemented in cTAKES an SVM-based approach to identify disease severity and body location. 

The first machine learning approach that was applied to predict multiple modifiers for a preidentified clinical entity was by Xu et al. \citep{xu-etal-2015-uth} when the ShARe corpus \citep{elhadad-etal-2015-semeval} was released. The authors have systematically extracted multiple features, namely context of up to 8 words before and after an entity, dependency relation, section names, and other lexicon features to train an individual SVM classifier for each modifier. They achieve state-of-the-art (SOTA) results in identifying modifiers based on gold disorders.
More recently, researchers have applied deep learning techniques such as long short-term memory (LSTM) to detect modifiers of medical entities in a clinical text \citep{xu2019applying, shi2019extracting}. A bi-directional LSTM and a conditional random field (CRF) were used for Named Entity Recognition (NER) of modifiers by Xu et al \citep{xu2019applying}. They used randomly initialized word embedding and position embedding because pre-trained word embeddings like Glove did not show much improvement in their experiments. If a sentence has more than one entity, they generate multiple training samples from the same sentence – one for each target entity and label all the modifiers of that entity using the BIO scheme. The authors used accuracy as the metric to evaluate the performance of their model. While this novel approach overcomes the problem of omitted annotations, its performance is much lower than the previous SOTA model.

A multi-task approach of extracting entities and their modifiers was implemented by Shi et al. They trained a Bi-LSTM-CRF model to extract clinical entities and modifiers and Bi-LSTM to predict the relation between every entity and modifier within the clinical text. Additionally, they have applied some constraints to limit relations between entities and modifiers when a modifier cannot be applied to an entity  \citep{shi2019extracting}. 


The advent of transformer models \citep{Vaswani17} has revolutionized various aspects of NLP through the application of transfer learning. A recent contribution in this field is the work by Khandelwal and Britto \citep{khandelwal-britto-2020-multitask}, who innovatively utilized transformer-based models for predicting negation and speculation modifiers in a multitask learning framework. Their methodology involved the fine-tuning of three different pre-trained transformer encoders: BERT \citep{devlin2018bert}, XLNet\citep{yang2019xlnet}, and RoBERTa \citep{liu2021robustly}. This process was coupled with a unique approach of using a single classification head and injecting special tokens into the input text, guiding the model toward the specific task at hand—either negation or speculation detection. While their approach contributed to the broader understanding of multitask learning in NLP, their research did not extend to exploring these methods in the context of clinical texts, an area ripe for further investigation.

\subsection{Contribution}
In contrast to Khandelwal and Britto's \citep{khandelwal-britto-2020-multitask} approach, our research extends the application of transfer learning and multi-task training to the specific area of entity modifier classification within clinical texts. We apply transfer learning to two diverse data sets, outperforming existing benchmarks on a publicly available data set. Our methodology differs notably from that of Khandelwal and Britto; we do not rely on the injection of special tokens into our input data. Instead, we employ a distinct classification head for each modifier, with simultaneous training of all heads to enhance model efficiency. This study also showcases the versatility of multi-task training in transferring weights across modifier heads from different data sets, even with only partial modifier overlap. Additionally, through ablation studies, we identify the most effective architectural configurations for our approach. A practical real-world application of our method is demonstrated in addressing the identification of modifiers for clinical entities specific to Opioid Use Disorder (OUD).

\section{Methods} \label{Materials_Methods}

\subsection{Data} \label{data}
We utilized two data sets for the evaluation of modifier detection in clinical text. The published ShARe corpus from SemEval 2015 Task 14 \citep{elhadad-etal-2015-semeval} and an unpublished corpus from the University of Alabama at Birmingham that was annotated for OUD-related entities (including modifiers). An overview of both corpora, including the number of documents, entities, modifier types, and counts, is shown in Table~\ref{tab-ShARe-stat}.

\subsubsection{ShARe Data Set} \label{share}
Task 14 of the SemEval-2015 \citep{elhadad-etal-2015-semeval} provided a data set for two tasks, including clinical disorder name entity recognition and template slot filling. It consists of 531 de-identified clinical notes. For this publication, we focus only on the template slot filling task. This task requires the identification of negation, severity, course, subject, uncertainty, conditional, and generic modifiers of the clinical disease entity. 
The assigned training and development set are combined to build our final model, and the test set results are reported. 

\subsubsection{OUD Data Set} \label{OUD_corpus}
After training by WC (OUD research co-coordinator) anotators created a corpus consisting of 3295 clinical notes from 59 patients (23 controls) from physician case referrals between 2016 and 2021. Annotation of 25478 OUD entity mentions and modifiers were done using BRAT 1.3 software. Annotators modified entities for negation, subject and assigned a DocTime value of before, overlaps, or after. Additionally, annotators annotated mentions of substance and opioid use, OUD and Substance Use Disorder (SUD) as illicit. To the best of our knowledge, illicitDrugUse is a unique event modifier in our data set. 
We split the data set to 80\% for training, 10\% for development, and 10\% for testing based on entities, not documents. The training and development set are combined to build our final model, and the test set results are reported. We plan to de-identify and release this data set as part of a future shared task.

\begin{table}
\caption{Statistics of the ShARe and OUD Corpus}\label{tab-ShARe-stat}
\centering
\begin{tabular}{lllllllllll}
\hline
\textbf{Corpus} & \textbf{Ents}  & \textbf{Sev}  & \textbf{Cou}	& \textbf{Unc} & \textbf{Con} & \textbf{Gen}  & \textbf{Neg} & \textbf{Sub} & \textbf{DT} & \textbf{IDU}\\
\hline
ShARe  & 25632 & 2067 & 1307 & 1848 & 1414 & 238 & 4932 & 248 & - & - \\

\hline
OUD & 28807 & 33 & - & 176 & - & - & 2523 & 487 & 2901 & 825 \\
\hline
\end{tabular}
\footnotetext{Default modifier values are not reported. Ents: entities, Neg: negation, Sev: severity,  Cou: course, Sub: subject, Unc: uncertainty,  Con: conditional, Gen: generic, DT: DocTime, IDU: Illicit Drug Use.}
\end{table}

\subsection{Architecture} \label{base_model}
\begin{figure*}[ht]
    \centering
    \includegraphics[width=13cm, height=6.5cm,trim={0 0cm 0 0cm},clip]{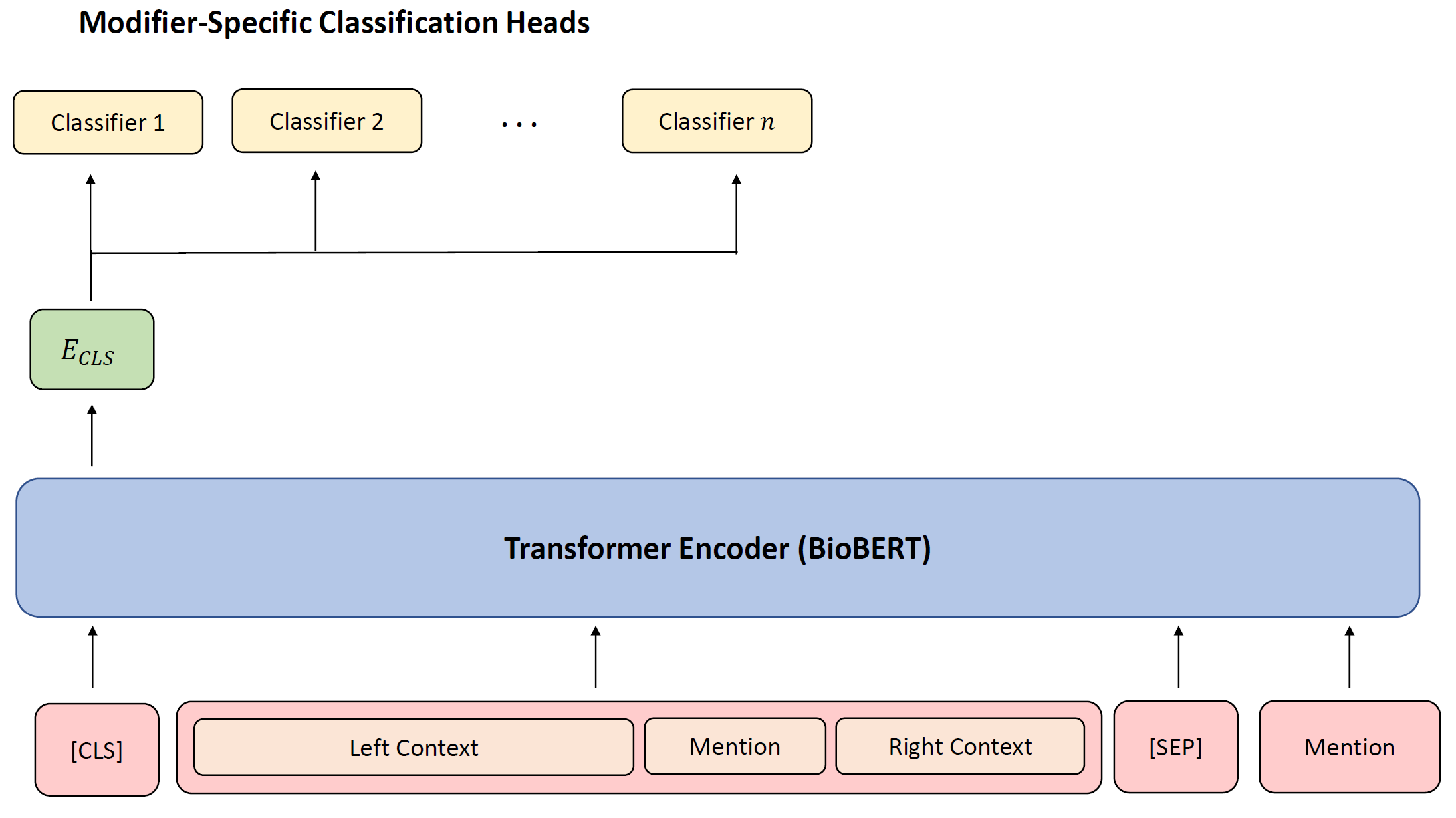}
    \caption{Overview our modifier predication model. The multi-task architecture contains a Classification head for each distinct modifier type. The single-task architecture has only a single head for the classification of one of the modifiers.}
    \label{fig:model}
\end{figure*}

We created a single-task and a multi-task architecture as shown in Figure \ref{fig:model}, where the single-task architecture uses a single classification head for all modifiers whereas the multi-task architecture has a head for each clinical modifier. Both architectures use BioBERT \citep{lee2020biobert} as the base model. We chose BioBERT over other variants like BERT \cite{devlin2018bert}, ClinicalBERT \citep{alsentzer2019publicly}, and PubMedBERT \citep{gu2021domain} since it performed better for detecting modifiers at initial experiments. No additional pre-training is performed using text from either the ShaRe Corpus or OUD corpus. 

\subsubsection{Models}
\begin{figure*}[ht]
    \includegraphics[width=12.05cm, height=5.65cm,clip]{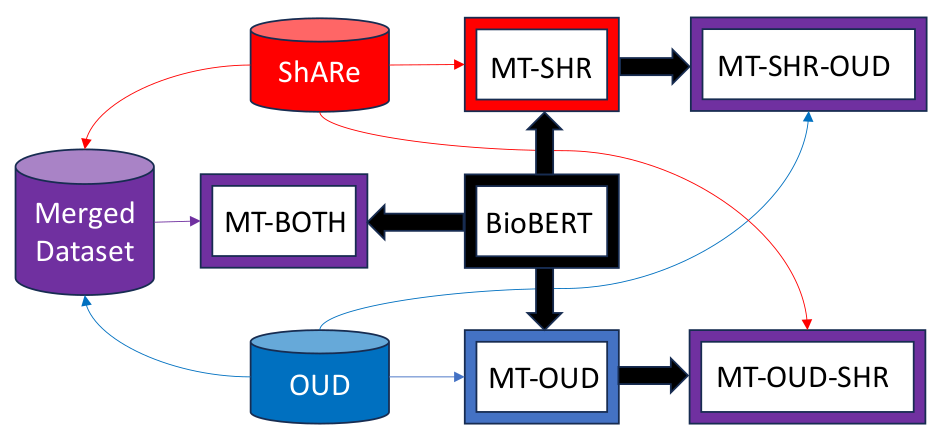}
    \caption{Overview of transfer learning. Thin arrows indicate training data flow, color-coded for the data source. Thick arrows indicate fine-tuning operations.}
    \label{fig:model_training}
\end{figure*}

 We name our single-task fine-tuned BioBERT model \textbf{ST}, our multi-task fine-tuned model \textbf{MT}. For our transfer learning experiments, models have a dash separated suffix indicating which data set they were fine-tuned on. For example,  \textbf{SHR} or \textbf{OUD} reference fine-tuning on the ShARe corpus or OUD corpus respectively. Fine-tuning operations are ordered from left (most distant) to right (most recent) based on the order in which they occurred. See Figure \ref{fig:model_training} for reference.
The \textbf{MT-SHR-OUD}  and \textbf{MT-OUD-SHR} have 5 and 7 classification heads, respectively. We also perform an experiment by combining the two data sets called \textbf{MT-BOTH} with a total of 9 distinct classification heads representing all clinical modifiers from both data sets.

For all models we use the final hidden vector corresponding to the [CLS] token (\begin{math} \mathbf{h_{cls}} \in \mathbb{R}^H \end{math}) generated from BioBERT($\widehat{X}$) as the common feature vector that is passed to each modifier classification head, which is a linear layer with learnable parameter \begin{math} \mathbf{W}_i \in \mathbb{R}^{m_i \times H} \end{math}. Formally, the probability distribution of a modifier type is:


\begin{equation}\label{prob}
    P(
    P(\hat{y}_{m_i}|\widehat{X}) = \mathrm{softmax}(\mathbf{W}_i \mathbf{h}_{cls} + \mathbf{b}_i)
\end{equation}
where $y_{m_i}$ is a label of the modifier $m_i$. We train the model using cross-entropy loss for each classifier:
\begin{equation}\label{loss}
     L_i = \sum_{1}^{k} P(y_{m_i}) \log P(\hat{y}_{m_i}|\widehat{X})
\end{equation}
where $k$ is the length of the batch. The final loss is the average of all classifiers. \\
\begin{equation}
    L = 1/n \sum_{1}^{n} L_i
\end{equation}
where $n$ represents the number of modifier heads in the model. Additionally, we experiment with the use of focal loss \citep{lin2017focal} on our model, a type of loss function commonly used in deep learning for tasks involving imbalanced data sets.

\subsubsection{Feature Extraction} 
We have adopted the question-answering input format described in the original BERT \cite{devlin2018bert} to fine-tune BioBERT and adapt it to modifier prediction.
As illustrated in Figure ~\ref{fig:model}, the model gets two sequences as input. The \(left context\) sequence is a disorder mention and its context. We chose the context to be 200 characters before the mention and 50 characters after, an empirically driven hyper-parameter choice that achieved better performance than word-based and sentence-based context. We hypothesize this to the fact that sentence boundaries are not well-formed in clinical text \cite{griffis2016quantitative}. For instance, some clinical notes only use commas throughout whole paragraphs without any periods. The \(right context\) sequence is the string of the entity itself. If the entity is discontiguous, only the strings that represent the entity are used. An example of what is passed to the model is: \verb|[CLS]| \emph{ The patent[\it{sic}] was found to be in fulminant liver failure. There she was having hallucinations, suicidal ideations and …} \verb|[SEP]| \emph{hallucinations}. The second example would include the first sequence and \emph{suicidal ideations} as a second sequence. 


The goal of this design is to help direct the model's attention to focus on the desired entity to extract its modifiers. Specifically, the entity is contextualized with the surrounding words, which generally include the modifiers in the first sequence. In addition, the second sequence redirects the aggregate sequence representation \verb|[CLS]| attention to the entity under consideration (\emph{hallucinations} in the first aforementioned example).

\subsection{Model Training} 
For training, we follow standard procedures and use the curated training data set for both data sources to develop our models. Hyper-parameters are optimized using the designated development set. We have trained our model for 10 epochs with an empirically derived early stop. The maximum sequence length is 144, the learning rate is 2e-5, the weight decay is 1e-2, and the batch size is 64. AdamW is used as our optimizer. Similar to Xu et al. \citep{xu-etal-2015-uth}, the training and development set are combined to build our final model. We report the results on their respective test sets. We used a single Tesla P100 GPU with 16GB memory to run all experiments. The model will be made available upon the acceptance of the publication.

\subsection{Experiments}
We conducted the following experiments:
\begin{itemize}
  \item To assess the performance of transfer learning and multi-task training for clinical modifiers, we evaluate \textbf{MT} on the OUD corpus and the ShARe corpus. We compare our results to previously reported results for the ShARe corpus and against a generalized clinical modifier model that combines all training examples from both corpora.

    \item To evaluate the feasibility of domain adaptation for clinical modifiers when only a portion of clinical modifiers match the target and source domain,  we perform bidirectional fine-tuning between the OUD and ShARe corpus. We fine-tune on the source domain, then perform an additional round of fine-tuning on the target domain to create 2 models (\textbf{MT-SHR-OUD}) and (\textbf{MT-OUD-SHR}) that are fine-tuned first on the ShARe corpus or OUD corpus respectively. We include a classification head for each modifier from both data sets for a total of \# heads. 
    \item We performed 2 ablation studies on both the OUD and ShARe corpus by (1) removing the disorder mention after the \verb|[SEP]| or (2) replacing the MT model heads with a single-headed (\textbf{ST}) model, where a model is trained separately for each modifier. 
\end{itemize}
    
\subsection{Evaluation}

To evaluate a system for identifying rare values of different modifiers, the original challenge for the ShARe corpus used weighted accuracy, where the prevalence of different values for each of the modifiers is considered. Specifically, for each modifier \(m_i\) the weights are calculated as follows
\[weight (m_i^k )=1-prevalence (m_i^k)\]
where \(k\) represents the different classes of the modifier \(m_i\) as described in the task description paper \citep{elhadad-etal-2015-semeval}. We used the evaluation script from the challenge organizers to compare with the previous state-of-the-art system \citep{xu-etal-2015-uth}. We have also used the standard unweighted accuracy and micro-averaged F1 to compare against other later works (Table~\ref{tab-mod-results}). For the OUD data set, we used the standard unweighted accuracy and macro-averaged F1 (including the null class) and micro-averaged F1 for possible future comparison.

\subsubsection{Chi-square test}
We perform a Chi-square test \citep{pearson1900x} to compute the statistical difference between our model results and previous results through comparing the number of correct and incorrect predictions. We used accuracy and the total number of examples in the test set if these numbers were not available.

\section{Results}
Multi-task training and transfer learning performance are shown in Table~\ref{tab-mod-results} for the ShARe corpus and in Table~\ref{tab-OUD-results} for the OUD corpus. Detailed explanations of the results are in the next subsections. 


\begin{table}[ht]
\centering
\caption{\label{tab-mod-results}
Model Performance on the ShARe corpus
}
\begin{tabular}{llllllllll}
\hline
Model  & Ep & Cou & Sub & Gen & Con & Unc & Neg & Sev & Avg\\
\hline
\multicolumn{10}{c} {weighted accuracy} \\
\hline
cTAKES 4.0.0.1 & - &  \hspace{.2cm} - & 0.793  & 0.828 & 0.485 & 0.587 & 0.715 & \hspace{.3cm} - &  \hspace{.3cm} - \\
SOTA \citep{xu-etal-2015-uth} & - &  0.887  & 0.975  & 0.911 & \textbf{0.903}  & 0.911  & 0.975  & 0.936 & 0.928 \\
MT-SHR & 5 &  \underline{0.922} & \textbf{\underline{0.980}} & \textbf{\underline{0.924}} & 0.873  & \textbf{\underline{0.950}} & \textbf{0.977} & \textbf{\underline{0.948}} & \textbf{0.939} \\
MT-SHR.fl  & 5 &  \textbf{0.927}  & 0.949  & 0.866 & 0.865  & 0.909  & 0.966  & \textbf{0.950} & 0.919\\
\hline

\multicolumn{10}{c} {micro average F1} \\
\hline
Shi et al. \citep{shi2019extracting} & - &  0.591 & 0.737 & 0.212 & 0.546 & 0.440 & 0.771 & 0.774 & 0.581 \\
Xu et al. \citep{xu2019applying} & - &  0.677 & 0.722 & \hspace{.2cm} - & 0.710 & 0.461 & 0.823 & 0.830 &  0.704 \\
MT-SHR & 5 &  \textbf{\underline{0.807}} & \underline{0.840}* & \underline{0.519} & \underline{0.818} & \textbf{\underline{0.861}} & \underline{0.954} & \textbf{\underline{0.884}} & 0.813\\
MT-BOTH & 20 &  0.796 &  \underline{0.918}* & 0.308 & 0.813 & 0.844 & \underline{0.965} & 0.867 & 0.787 \\
MT-OUD-SHR & 5 &  0.791 & \textbf{\underline{0.958}}* & \textbf{\underline{0.630}} & \textbf{\underline{0.831}} & 0.850 & \textbf{\underline{0.977}} & \textbf{0.884} & \textbf{0.846}\\
\hline

\multicolumn{10}{c} {unweighted accuracy} \\
\hline
Xu et al. \citep{xu2019applying} & - &  0.970 & 0.994 & \hspace{.2cm} - & 0.972 & 0.955 & 0.954 & 0.982  &  0.971\\
MT-SHR & 5 &  \textbf{\underline{0.984}} &  \textbf{0.998} & 0.993 & \underline{0.981} & \textbf{\underline{0.982}} & \textbf{\underline{0.992}} & \textbf{0.987} & \textbf{0.988} \\
MT-BOTH & 20 & \textbf{0.984} &  \textbf{0.998} & 0.976 & 0.967 & 0.981 & 0.987 & 0.983 & 0.982\\
MT-OUD-SHR & 5 & 0.983 &  \textbf{0.998} & \textbf{0.994} & \textbf{0.982} & \textbf{0.982} & \textbf{0.992} & 0.984 & \textbf{0.988} \\
\hline
\end{tabular}
\footnotetext{ \textbf{Bold} font means the best performance. \underline{Underline} means statistically significant \emph{p-value} relative to the model in the previous line. Ep: epochs, Neg: negation, Sev: severity, Cou: course, Sub: subject, Unc: uncertainty, Con: conditional, Gen: generic, Avg: average, fl: focal loss. * For micro averages of MT, MT-BOTH and MT-OUD-SHR, the subject modifier the 'other' class is excluded due to having only 4 examples.}
\end{table}

\subsection{Comparative Performance of Multi-Task Training}
We compare our results to previously reported results using the same metrics as the originally reported result. For the ShARe corpus, the top portion of Table~\ref{tab-mod-results} contains the results from using the rule-based cTAKES 4.0.0.1 system, the previous SOTA model \citep{xu-etal-2015-uth} and our multi-task (MT-SHR) model on the ShARe corpus using the weighted accuracy measure. We also report for micro average F1 score and unweighted accuracy in the middle and bottom portion of Table~\ref{tab-mod-results} respectively. Our multi-task model (MT-SHR) performs better in all modifiers except the conditional modifier based on weighted accuracy and is on par with the SOTA model for the negation modifier. Our model shows a statistically significant improvement in all other modifiers by at least 1\%. For example, the uncertainty modifier scored the highest improvement of 4\% among other modifiers. On average, our model improved the performance by slightly over 1\%. The low performance on the conditional modifier had a disproportionate impact on the final average results. In both corpora, training with cross-entropy loss was superior to focal loss despite class imbalance in modifier distribution.

\subsection{Domain Adaptation}
Results for domain adaption are shown in Table~\ref{tab-mod-results} for the ShARe corpus (MT-OUD-SHR) and in Table~\ref{tab-OUD-results} for the OUD corpus (MT-SHR-OUD). 
Initial training on one data set followed by additional training on another data set (MT-OUD-SHR in Table~\ref{tab-mod-results} and MT-SHR-OUD in Table~\ref{tab-OUD-results}) with partially similar modifiers led to at least similar or even better performance compared to training on a single data set. On micro average F1 score, the performance improved by 3.3\% compared to training only on the ShARe data set. However, merging the two data sets (MT-BOTH) decreased the average performance by 2.6\%. 

\begin{table}[ht]
\centering
\caption{\label{tab-OUD-results}
 Model Performance on the OUD Corpus Modifiers. }

\begin{tabular}{llllllll}
\hline
Model & Ep & Neg & Sub & DT  & IDU & Unc & Sev \\
\hline
\multicolumn{8}{c} {macro average F1} \\
\hline
MT-SHR & \hspace{.1cm} - & 0.769 & 0.863 & \hspace{.3cm} - & \hspace{.3cm} - & 0.542 & 0.547 \\
MT-OUD & \hspace{.1cm} 3 &  \textbf{\underline{0.958}} & \textbf{\underline{0.920}} & 0.838 & 0.839 & \textbf{\underline{0.833}} & \hspace{.3cm} - \\
MT-OUD.fl & \hspace{.1cm} 4 & 0.892 & 0.855 & 0.816 & 0.784 & 0.632 & \hspace{.3cm} - \\
MT-BOTH & \hspace{.1cm} 20 & 0.951 & 0.911 & 0.846 & \textbf{\underline{0.867}} & 0.765 & \textbf{\underline{0.929}} \\
MT-SHR-OUD & \hspace{.1cm} 6 & 0.948 & 0.915 & \textbf{\underline{0.877}} & 0.848 & 0.772 &  \hspace{.3cm} - \\
\hline
\multicolumn{8}{c} {micro average F1} \\
\hline
MT-SHR & \hspace{.1cm} - & 0.591 & 0.732 & \hspace{.3cm} - & \hspace{.3cm} - & 0.100 & 0.105 \\
MT-OUD & \hspace{.1cm} 3 &  \textbf{\underline{0.912}} & \textbf{\underline{0.842}} & 0.776 & 0.688 & {\underline{0.546}} & \hspace{.3cm} - \\
MT-OUD.fl & \hspace{.1cm} 4 & 0.803 & 0.714 & 0.729 & 0.583 & 0.267 & \hspace{.3cm} - \\
MT-BOTH & \hspace{.1cm} 20 & \textbf{0.912} & 0.825 & 0.781 & \textbf{\underline{0.743}} & 0.533 & \textbf{\underline{1.00}} \\
MT-SHR-OUD & \hspace{.1cm} 6 & \textbf{0.912} & \textbf{0.842} & \textbf{\underline{0.816}} & 0.705 & \textbf{\underline{0.593}} & \hspace{.3cm} - \\
\hline
\multicolumn{8}{c} {unweighted accuracy} \\
\hline
MT-SHR & \hspace{.2cm} - & 0.908 & 0.987 & \hspace{.3cm} - & \hspace{.3cm} - & 0.968 & 0.979 \\
MT-OUD & \hspace{.2cm} 3 &  \textbf{\underline{0.984}} & \textbf{0.995} & 0.953 & 0.980  & \textbf{\underline{0.997}} & \hspace{.3cm} - \\
MT-OUD.fl & \hspace{.2cm} 4 & 0.965 & 0.992 & 0.941 & 0.969 & 0.995 & \hspace{.3cm} - \\
MT-BOTH & \hspace{.1cm} 20 &  \textbf{0.984} & 0.994 & 0.954 & \textbf{0.984} & 0.995 & \textbf{0.982} \\
MT-SHR-OUD & \hspace{.2cm} 6 & 0.983 & 0.994 & \textbf{0.960} & 0.982 & 0.996 &  \hspace{.3cm} -  \\
\hline
\end{tabular}
\footnotetext{\textbf{Bold} font means the best performance. \underline{Underline} means statistically significant \emph{p-value} relative to the model in the previous line. Ep: epochs, Neg: negation, Sev: severity, Sub: subject, Unc: uncertainty, DT: DocTime, IDU: Illicit Drug Use}
\end{table}

\subsection{Ablation Study}
Results for our ablation studies are shown for the ShARE corpus and OUD codrpus in Table~\ref{tab-ablation-table} and Table~\ref{tab-OUD-ablation} respectively. Removal of the mention after the \verb|[SEP]| or "no hint" decreases the performance of the model for both data sets and has been seen in similar research experiments ~\citep{webson-pavlick-2022-prompt}.
The second to the last row in Table~\ref{tab-ablation-table} and Table~\ref{tab-OUD-ablation} show results evaluating the effectiveness of multi-task training \textbf{MT} versus standard fine-tuning \textbf{ST} on each modifier separately. For the ShARe data set, the performance dropped by an average of 2\% compared with MT-SHR. It is even worse than the previous SOTA, which is based on SVM by 1\% on average.

\begin{table}[ht]
\centering
\caption{\label{tab-ablation-table}
Ablation study results on the ShARe Corpus }
\begin{tabular}{llllllllll}
\hline
Model & Ep & Cou & Sub & Gen  & Con  & Unc & Neg & Sev & Avg\\
\hline
\multicolumn{10}{c} {weighted accuracy} \\
\hline
SOTA \citep{xu-etal-2015-uth} & - & 0.887  & 0.975  & 0.911 & \textbf{0.903}  & 0.911  & 0.975  & 0.936  & 0.928 \\
MT-SHR & 5 & \textbf{0.922} & \textbf{0.98} & \textbf{0.924} & 0.873  & \textbf{0.95} & \textbf{0.977} & \textbf{0.948} & \textbf{0.939} \\
ST-SHR & 2,3 & 0.912 &  0.959  & 0.873 & 0.849  & 0.938  & 0.973  & 0.915  & 0.917 \\
MT-SHR-no hint & 5 & 0.880 & 0.929 & 0.872 & 0.874 & 0.890 & 0.955 & 0.913 & 0.902 \\
\hline
\end{tabular}
\footnotetext{Ep: epochs, Neg: negation, Sev: severity, Cou: course, Sub: subject, Unc: uncertainty, Con: conditional, Gen: generic, Avg: average.}
\end{table}

\begin{table}[ht]
\caption{\label{tab-OUD-ablation}
 Ablation study results for the OUD corpus.}
\begin{tabular}{lllllll}
\hline
Model & Ep & Neg & Sub & DT  & IDU & Unc \\
\hline
\multicolumn{7}{c} {macro average F1} \\
\hline
MT-OUD & 3 & \textbf{0.958} & \textbf{0.920} & 0.838 & \textbf{0.839}	& \textbf{0.833}  \\
ST-OUD & 2,3 & 0.935 & 0.911 & \textbf{0.846} & 0.831 & 0.768  \\
MT-OUD-no hint & 3 & 0.912 & 0.880 & 0.739 & 0.689 & 0.721 \\

\hline
\multicolumn{7}{c} {unweighted accuracy} \\
\hline
MT-OUD & 3 & \textbf{0.984} & \textbf{0.995} & \textbf{0.953} & 0.980  & \textbf{0.997} \\
ST-OUD & 2,3 & 0.971 &  0.993  & 0.928 & \textbf{0.982} & 0.996  \\
MT-OUD-no hint & 3 & 0.945  & 0.988 & 0.910 & 0.968 & 0.995 \\
\hline
\end{tabular}
\footnotetext{Severity was removed due to too few examples. Ep: epochs, Neg: negation, Sub: subject, Unc: uncertainty, DT: DocTime, IDU: Illicit Drug Use}
\end{table}

\section{Discussion}
\paragraph{Comparison to Previous Work}
The adoption of transfer learning and multi-task training (MT) yields an overall improvement of 10 points (MT-SHR) on the micro-F1 score when compared to previous work by Xu et al.\citep{xu2019applying} using a Bi-LSTM architecture as shown in Table \ref{tab-mod-results}. This improvement rises to 14 points when an initial round of fine-tuning is done on the OUD corpora (MT-OUD-SHR). This increase occurs even when there is only partial alignment of modifier types between the two data sets. Combining multiple data sets (MT-BOTH), similar to the work of Khandelwal and Britto work \citep{khandelwal-britto-2020-multitask} yields a model that can predict all modifiers from both input data sets at the expense of performance.

\paragraph{Transfer Learning on Modifiers Common between Data Sets}
Entity modifiers for negation, subject, severity and uncertainty are shared between the ShARe and OUD data sets. For these commons entity modifiers, our results indicate that the target data set can benefit from a previous round of fine-tuning on the source data set. This benefit is more pronounced when the set of modifiers shared between datasets have sufficient training instances. For example, micro F1 performance was improved on the ShARe corpus for the shared negation and subject modifiers, both of which have a proportionally higher number of training examples. We are uncertain as to why the same benefit was not shown on the OUD corpus for these high frequency common modifiers, but the additional round of fine-tuning did not decrease performance. For low frequency modifiers like the uncertainty modifier, results improved when transferring from ShARe to OUD (MT-SHR-OUD) but decreased when transferring from OUD to ShARe (MT-OUD-SHR) compared to MT since OUD has few examples with annotated uncertainty. Severity modifier instances are too low frequency in the OUD corpus to be evaluated for this transfer learning experiment, so we show performance on the ShARe and combined corpus only.

\paragraph{Transfer Learning on Modifiers Uncommon between Data Sets}
One surprising result of our approach is that transfer learning performance was improved on modifiers found only in the target data set, even when they were not present in the source data set. We can see this for the result of course, generic, and conditional in MT-OUD-SHR and DT and IUD in MT-SHR-OUD. This could be the result of the initial round of training where the model has seen more examples to learn the task. However, combining the two data sets for training one model (MT-BOTH) had mixed effects on the two data sets. This may the result of relative class imbalance for entity modifier types between source and target data sets.

\paragraph{Efficiency of Transfer Learning between Data Sets} 
Multi-task training gave better performance compared to single-task training (see Table~\ref{tab-ablation-table}) and it was more efficient. On average, it took 6 minutes to train our MT model for one epoch, while it took 5 minutes to train an ST model, which had to be repeated for each modifier. Overall, the training time and resources cost for MT model was reduced by at least 60\% compared to training all ST models. Multi-task training and transfer learning from one data set to the other was also more efficient than combining the two data sets (MT-BOTH). Training MT-BOTH required 20 epochs for both data sets whereas two consecutive rounds of multitask transfer learning (MT-OUD-SHR and MT-SHR-OUD) reduced that cost by almost 50\%. We noticed that MT-BOTH needed significantly more time to learn the uncommon modifiers compared to learning the common modifiers. While combining the data sets is less efficient, it allowed MT-BOTH to overcome the problem of significantly low examples in a data set, such as the OUD data set severity modifier.

\subsection{Error Analysis}
In our study, we carefully selected 10 errors from each modifier type for a total of 70 errors for the ShARe and 50 for the OUD test sets. This gave us a diverse range of errors to examine closely. We used the prediction of MT-SHR and MT-OUD for our analysis. A key finding from our analysis was that in 44\% of cases from the ShARe, our model made correct predictions, but these appear to be have been annotated incorrectly in the data sets. For example, in the sentence \emph{"There was no rebound or guarding,"} our model correctly identified the negation, but this was not reflected in the gold data set. Another instance involved a discrepancy in severity grading: our model classified a case as \emph{moderate}, while the data set labeled it as \emph{severe} for the example \emph{"Mild degenerative changes are seen throughout the spine"}. This kind of inconsistency was also present in the initial version of the OUD dataset, but we have since rectified these issues, elevating its reliability to at least match, if not exceed, that of the ShARe data set.

In the context of modifier evaluations, false negatives occur when the system predicts the  'null', 'unmarked' or default class. This occurred in 46\% of our examples in the ShARe dataset and 58\% in the OUD data set. The default case covers an average of 92\% of instances in both data sets, this class imbalance means our model tends to overlook or incorrectly classify under-represented modifiers. For instance, in \emph{"He had slight decreased sensation in his right upper extremity,"} the model failed to identify \emph{'slight'} as the severity level, opting instead for \emph{'unmarked'}. Within the OUD data set, the model struggles to identify the negation present in the clinical note section for the psychological state. It also struggles with abbreviations or alternatives for negations, like ‘NEG’, ‘0’, 'Zero’, and ‘None’. For instance, \emph{Family History: 0 suicide attempts} has a \emph{0} that was not recognized by the model as a negation for \emph{suicide attempts}. For the DocTime modifier, the model has some trouble recognizing the current annotated text as a part of the history section of the clinical note – deeming all information included as ‘Before’.

False positives accounted for about 54\% of ShARe and 42\% for OUD errors. These errors often stemmed from confusing contexts. For example, in the phrase  \emph{"Per her daughter she has been having shortness of breath,"} the model misinterpreted  \emph{'shortness of breath'} as referring to the daughter, not the patient. Similarly, in the OUD dataset, the model was confused by drug/lab test contexts. In sentences like  \emph{"Lab Results: U Methadone Negative U Opiates Positive U Oxycodone Negative,"} the presence of the word  \emph{'Negative'} misled the model for the affirmed mention \emph{'U Opiates'}.

Lastly, our system made unexpected errors about 10\% of the time. On occasion, it failed to recognize terms such as "denies" as negations. This may reflect the varied contextual language and inconsistent use of denies by physicians where physicians suspected but patient denied conditions are marked as denies whereas annotation guidelines impose consistency on a more nuanced note. Other errors were due to fine distinctions between different classifications, such as failing to differentiate between 'increased' and 'worsened' in symptom descriptions. For instance, the model predicted that the \emph{fatigue} increased while it was annotated as worsened in the example \emph{"Three days of progressive fatigue.”} 

For more examples and a detailed look at these points, please refer to Table~\ref{tab-error-analysis}.
\begin{table}[ht!]
\centering
\caption{\label{tab-error-analysis} TEST Error Analysis for the ShARe and OUD Data Set.}

\begin{tabular}{V{1.2cm}|p{2.6cm}|p{2.6cm}|p{2.65cm}|c}
\toprule
Error Type & \multicolumn{3}{c|}{Example} & Proportion \\
 & MODIFIER & ANNOTATION & MODEL &  \\
\midrule




\multirow{6}*{\parbox{1.2cm}{False negative (ShARe)}}  & 
\multicolumn{3}{V{9cm}|}{\textbf{Ex1:} His extremities also showed greater \underline{swelling in his left leg}.} & \multirow{6}*{ 46\%}  \\
& COURSE &  INCREASED & UNMARKED & \\\cmidrule{2-4}

 & 
\multicolumn{3}{V{9cm}|}{\textbf{Ex2:} He had slight decreased \underline{sensation} in his right upper extremity.} & \\
 & SEVERITY & SLIGHT & UNMARKED & \\\cmidrule{2-4}

 &
 \multicolumn{3}{V{9cm}|}{\textbf{Ex3:} \underline{Peptic ulcer disease} in the OMR, but the patient denies this. } & \\
& NEGATION & YES & NO & \\ 

\midrule
\multirow{6}*{\parbox{1.2cm}{False positive (ShARe)}}  & 
\multicolumn{3}{V{9cm}|}{\textbf{Ex1:} These findings are consistent with a \underline{hematoma} which appears more organized than on prior exam.} & \multirow{6}*{ 54\%}  \\
& UNCERTAINTY &  NO & YES & \\\cmidrule{2-4}
 & 
\multicolumn{3}{V{9cm}|}{\textbf{Ex2:} Per her daughter she has been having \underline{shortness of breath}.} & \\
 & SUBJECT & PATIENT & FAMILY MEMBER & \\\cmidrule{2-4}
 &
\multicolumn{3}{V{9cm}|}{\textbf{Ex3:} Family History: Father died 66 from heart failure Mother died 59 from cervical cancer. Diabetes in father's family as well as \underline{heart disease}.} & \\
& SUBJECT & OTHER & FAMILY MEMBER & \\

\midrule
\midrule
\multirow{6}*{\parbox{1.2cm}{False negative (OUD)}}  & 
\multicolumn{3}{V{9cm}|}{\textbf{Ex1:} He was hospitalized in the CPM in 2011 for \underline{suicidal ideations} for several days.} & \multirow{6}*{ 58\%}  \\
& DOCTIME &  BEFORE & OVERLAP & \\\cmidrule{2-4}

 & 
\multicolumn{3}{V{9cm}|}{\textbf{Ex2:} Family History: 0 \underline{suicide attempts}.} & \\
 & NEGATION & YES & NO & \\\cmidrule{2-4}

 &
\multicolumn{3}{V{9cm}|}{\textbf{Ex3:} During interview she stated she has not been using illicit substances since D/C, but after UDS came back pos for \underline{amphetamines and benzos}, she admitted to using these illicitly about a week ago due to her high anxiety.} & \\
& IDU & TRUE & FALSE & \\ 
\midrule
\multirow{6}*{\parbox{1.2cm}{False positive (OUD)}}  & 
\multicolumn{3}{V{9cm}|}{\textbf{Ex1:} He said he was \underline{snorting heroin} and is not sure exactly how much he took.} & \multirow{6}*{ 42\%}  \\
& UNCERTAINTY & NO & YES & \\\cmidrule{2-4}
 & 
\multicolumn{3}{V{9cm}|}{\textbf{Ex2:} How has addiction impacted your relationships?: Was dating my husband when I was \underline{using suboxone}, hid it from him.} & \\
 & IDU & FALSE & TRUE & \\\cmidrule{2-4}
 &
\multicolumn{3}{V{9cm}|}{\textbf{Ex3:} Lab Results : U Methadone Negative \underline{U Opiates} Positive U Oxycodone Negative.} & \\
& NEGATION & NO & YES & \\ 

\bottomrule
\end{tabular}
\footnotetext{Mention in each example is \underline{underlined}. Longer context is ignored for the space limit. False negatives in this context are when the default modifier class is predicted. False positives are when the wrong modifier class is predicted. }
\end{table}

\subsection{Limitations}
To date the only corpus containing clinical modifiers of entities that has been published to our knowledge is the ShARe corpus from SemEval 2015 Task 14 \citep{elhadad-etal-2015-semeval}. This data set, in conjunction with the OUD data set, leaves only two data sets for evaluation. We do not evaluate large language models (LLMs) in this work, but do not believe this is needed, given our task is an information extraction task, fine-tuning was done and recent work suggests that domain models are capable of outperforming LLMs in this domain \citep{lehman2023we}. Additionally, we do not evaluate the anatomy modifier of the ShARe corpus since our approach requires training data for all class types and more than one-third of the classes in the ShARe test set are not in the training set. We are exploring synthetic data to address this issue. Finally, for the ShARe corpus, the vast majority of the examples have a single clinical entity mention within the chosen context. However, a duplicate clinical entity mention occurs in the same context window in 4\% of the examples and in 2\% of the examples the same clinical entity mention occurs 3 or more times. This can cause ambiguity since the clinical entity modifiers can only be distinguished by slightly different ends to their context window. 

\section{Conclusion}
Our results indicate that multi-task training can be beneficial for modifier identification and we show state-of-the-art performance on the ShARe corpus. Additionally, our experiments suggest that an additional round of fine-tuning on a similar data set can be more effective/efficient than training a transformer model on a combined data set, even if modifiers from the two data sets only partially overlap.


\section*{Declarations}
\begin{itemize}

\item {\bf Ethics approval and consent to participate:} Consent to participate was waived under IRB-121114001, "Using Text Mining to Extract Information from Text Documents in the Electronic Health Record". 
\item {\bf Consent for publication:} Not applicable, no personal details shared.
\item {\bf Competing interests:} The authors declare that they have no competing interests. 
\item {\bf Funding:} Funding for this work was provided by the Alabama Department of Mental Health OUD Center of Excellence. Work was done under IRB-300002304, "Initiating Medication Assisted Treatment of Opioid Addiction in the Emergency Department: The ED MAT Protocol" and IRB-121114001, "Using Text Mining to Extract Information from Text Documents in the Electronic Health Record".
\item {\bf Authors' contributions:} OUD project aspects and modifiers were conceived by LW, EE, and JO. The application of transfer and multi-task learning methods to these data sets was conceived by AIA and JO, who also wrote the original draft of the manuscript with AIA. Substantial edits were performed by SF, EE and WB. Project oversight was provided by JO and SF. Source code development was done by AIA and AA, overseen by JO. WC, CC, EC, ZD, and JH performed annotation work and revised annotation guidelines, with help from AIA, WB, and JO who were responsible for revisions. Error analysis was conducted by AIA, WC, and JH. All authors reviewed and approved the final manuscript.
\item {\bf Acknowledgements:} We would like to acknowledge Tobias O'Leary for OUD data set management and  UAB Research Computing for use of their hardware for these experiments. 
\end{itemize}

\bibliography{sn-bibliography}


\begin{thebibliography}{28}
\ifx \bisbn   \undefined \def \bisbn  #1{ISBN #1}\fi
\ifx \binits  \undefined \def \binits#1{#1}\fi
\ifx \bauthor  \undefined \def \bauthor#1{#1}\fi
\ifx \batitle  \undefined \def \batitle#1{#1}\fi
\ifx \bjtitle  \undefined \def \bjtitle#1{#1}\fi
\ifx \bvolume  \undefined \def \bvolume#1{\textbf{#1}}\fi
\ifx \byear  \undefined \def \byear#1{#1}\fi
\ifx \bissue  \undefined \def \bissue#1{#1}\fi
\ifx \bfpage  \undefined \def \bfpage#1{#1}\fi
\ifx \blpage  \undefined \def \blpage #1{#1}\fi
\ifx \burl  \undefined \def \burl#1{\textsf{#1}}\fi
\ifx \doiurl  \undefined \def \doiurl#1{\url{https://doi.org/#1}}\fi
\ifx \betal  \undefined \def \betal{\textit{et al.}}\fi
\ifx \binstitute  \undefined \def \binstitute#1{#1}\fi
\ifx \binstitutionaled  \undefined \def \binstitutionaled#1{#1}\fi
\ifx \bctitle  \undefined \def \bctitle#1{#1}\fi
\ifx \beditor  \undefined \def \beditor#1{#1}\fi
\ifx \bpublisher  \undefined \def \bpublisher#1{#1}\fi
\ifx \bbtitle  \undefined \def \bbtitle#1{#1}\fi
\ifx \bedition  \undefined \def \bedition#1{#1}\fi
\ifx \bseriesno  \undefined \def \bseriesno#1{#1}\fi
\ifx \blocation  \undefined \def \blocation#1{#1}\fi
\ifx \bsertitle  \undefined \def \bsertitle#1{#1}\fi
\ifx \bsnm \undefined \def \bsnm#1{#1}\fi
\ifx \bsuffix \undefined \def \bsuffix#1{#1}\fi
\ifx \bparticle \undefined \def \bparticle#1{#1}\fi
\ifx \barticle \undefined \def \barticle#1{#1}\fi
\bibcommenthead
\ifx \bconfdate \undefined \def \bconfdate #1{#1}\fi
\ifx \botherref \undefined \def \botherref #1{#1}\fi
\ifx \url \undefined \def \url#1{\textsf{#1}}\fi
\ifx \bchapter \undefined \def \bchapter#1{#1}\fi
\ifx \bbook \undefined \def \bbook#1{#1}\fi
\ifx \bcomment \undefined \def \bcomment#1{#1}\fi
\ifx \oauthor \undefined \def \oauthor#1{#1}\fi
\ifx \citeauthoryear \undefined \def \citeauthoryear#1{#1}\fi
\ifx \endbibitem  \undefined \def \endbibitem {}\fi
\ifx \bconflocation  \undefined \def \bconflocation#1{#1}\fi
\ifx \arxivurl  \undefined \def \arxivurl#1{\textsf{#1}}\fi
\csname PreBibitemsHook\endcsname

\bibitem[\protect\citeauthoryear{Chapman et~al.}{2001a}]{chapman2001evaluation}
\begin{bchapter}
\bauthor{\bsnm{Chapman}, \binits{W.W.}},
\bauthor{\bsnm{Bridewell}, \binits{W.}},
\bauthor{\bsnm{Hanbury}, \binits{P.}},
\bauthor{\bsnm{Cooper}, \binits{G.F.}},
\bauthor{\bsnm{Buchanan}, \binits{B.G.}}:
\bctitle{Evaluation of negation phrases in narrative clinical reports.}
In: \bbtitle{Proceedings of the AMIA Symposium},
p. \bfpage{105}
(\byear{2001}).
\bcomment{American Medical Informatics Association}
\end{bchapter}
\endbibitem

\bibitem[\protect\citeauthoryear{Chapman et~al.}{2001b}]{chapman2001simple}
\begin{barticle}
\bauthor{\bsnm{Chapman}, \binits{W.W.}},
\bauthor{\bsnm{Bridewell}, \binits{W.}},
\bauthor{\bsnm{Hanbury}, \binits{P.}},
\bauthor{\bsnm{Cooper}, \binits{G.F.}},
\bauthor{\bsnm{Buchanan}, \binits{B.G.}}:
\batitle{A simple algorithm for identifying negated findings and diseases in discharge summaries}.
\bjtitle{Journal of biomedical informatics}
\bvolume{34}(\bissue{5}),
\bfpage{301}--\blpage{310}
(\byear{2001})
\end{barticle}
\endbibitem

\bibitem[\protect\citeauthoryear{Chapman et~al.}{2013}]{chapman2013extending}
\begin{barticle}
\bauthor{\bsnm{Chapman}, \binits{W.W.}},
\bauthor{\bsnm{Hilert}, \binits{D.}},
\bauthor{\bsnm{Velupillai}, \binits{S.}},
\bauthor{\bsnm{Kvist}, \binits{M.}},
\bauthor{\bsnm{Skeppstedt}, \binits{M.}},
\bauthor{\bsnm{Chapman}, \binits{B.E.}},
\bauthor{\bsnm{Conway}, \binits{M.}},
\bauthor{\bsnm{Tharp}, \binits{M.}},
\bauthor{\bsnm{Mowery}, \binits{D.L.}},
\bauthor{\bsnm{Deleger}, \binits{L.}}:
\batitle{Extending the negex lexicon for multiple languages}.
\bjtitle{Studies in health technology and informatics}
\bvolume{192},
\bfpage{677}
(\byear{2013})
\end{barticle}
\endbibitem

\bibitem[\protect\citeauthoryear{Mirzapour et~al.}{2021}]{mirzapour2021french}
\begin{barticle}
\bauthor{\bsnm{Mirzapour}, \binits{M.}},
\bauthor{\bsnm{Abdaoui}, \binits{A.}},
\bauthor{\bsnm{Tchechmedjiev}, \binits{A.}},
\bauthor{\bsnm{Digan}, \binits{W.}},
\bauthor{\bsnm{Bringay}, \binits{S.}},
\bauthor{\bsnm{Jonquet}, \binits{C.}}:
\batitle{French fastcontext: A publicly accessible system for detecting negation, temporality and experiencer in french clinical notes}.
\bjtitle{Journal of Biomedical Informatics}
\bvolume{117},
\bfpage{103733}
(\byear{2021})
\end{barticle}
\endbibitem

\bibitem[\protect\citeauthoryear{Chapman et~al.}{2007}]{chapman2007context}
\begin{bchapter}
\bauthor{\bsnm{Chapman}, \binits{W.}},
\bauthor{\bsnm{Dowling}, \binits{J.}},
\bauthor{\bsnm{Chu}, \binits{D.}}:
\bctitle{Context: An algorithm for identifying contextual features from clinical text}.
In: \bbtitle{Biological, Translational, and Clinical Language Processing},
pp. \bfpage{81}--\blpage{88}
(\byear{2007})
\end{bchapter}
\endbibitem

\bibitem[\protect\citeauthoryear{Harkema et~al.}{2009}]{harkema2009context}
\begin{barticle}
\bauthor{\bsnm{Harkema}, \binits{H.}},
\bauthor{\bsnm{Dowling}, \binits{J.N.}},
\bauthor{\bsnm{Thornblade}, \binits{T.}},
\bauthor{\bsnm{Chapman}, \binits{W.W.}}:
\batitle{Context: an algorithm for determining negation, experiencer, and temporal status from clinical reports}.
\bjtitle{Journal of biomedical informatics}
\bvolume{42}(\bissue{5}),
\bfpage{839}--\blpage{851}
(\byear{2009})
\end{barticle}
\endbibitem

\bibitem[\protect\citeauthoryear{Shi and Hurdle}{2018}]{shi2018trie}
\begin{barticle}
\bauthor{\bsnm{Shi}, \binits{J.}},
\bauthor{\bsnm{Hurdle}, \binits{J.F.}}:
\batitle{Trie-based rule processing for clinical nlp: A use-case study of n-trie, making the context algorithm more efficient and scalable}.
\bjtitle{Journal of biomedical informatics}
\bvolume{85},
\bfpage{106}--\blpage{113}
(\byear{2018})
\end{barticle}
\endbibitem

\bibitem[\protect\citeauthoryear{Jagannatha et~al.}{2019}]{jagannatha2019overview}
\begin{barticle}
\bauthor{\bsnm{Jagannatha}, \binits{A.}},
\bauthor{\bsnm{Liu}, \binits{F.}},
\bauthor{\bsnm{Liu}, \binits{W.}},
\bauthor{\bsnm{Yu}, \binits{H.}}:
\batitle{Overview of the first natural language processing challenge for extracting medication, indication, and adverse drug events from electronic health record notes (made 1.0)}.
\bjtitle{Drug safety}
\bvolume{42},
\bfpage{99}--\blpage{111}
(\byear{2019})
\end{barticle}
\endbibitem

\bibitem[\protect\citeauthoryear{Elhadad et~al.}{2015}]{elhadad-etal-2015-semeval}
\begin{bchapter}
\bauthor{\bsnm{Elhadad}, \binits{N.}},
\bauthor{\bsnm{Pradhan}, \binits{S.}},
\bauthor{\bsnm{Gorman}, \binits{S.}},
\bauthor{\bsnm{Manandhar}, \binits{S.}},
\bauthor{\bsnm{Chapman}, \binits{W.}},
\bauthor{\bsnm{Savova}, \binits{G.}}:
\bctitle{{S}em{E}val-2015 task 14: Analysis of clinical text}.
In: \bbtitle{Proceedings of the 9th International Workshop on Semantic Evaluation ({S}em{E}val 2015)},
pp. \bfpage{303}--\blpage{310}.
\bpublisher{Association for Computational Linguistics},
\blocation{Denver, Colorado}
(\byear{2015}).
\doiurl{10.18653/v1/S15-2051} .
\burl{https://aclanthology.org/S15-2051}
\end{bchapter}
\endbibitem

\bibitem[\protect\citeauthoryear{Friedman et~al.}{1999}]{friedman1999natural}
\begin{barticle}
\bauthor{\bsnm{Friedman}, \binits{C.}},
\bauthor{\bsnm{Hripcsak}, \binits{G.}}, \betal:
\batitle{Natural language processing and its future in medicine}.
\bjtitle{Acad Med}
\bvolume{74}(\bissue{8}),
\bfpage{890}--\blpage{5}
(\byear{1999})
\end{barticle}
\endbibitem

\bibitem[\protect\citeauthoryear{Savova et~al.}{2010}]{Savova:2010hy}
\begin{barticle}
\bauthor{\bsnm{Savova}, \binits{G.K.}},
\bauthor{\bsnm{Masanz}, \binits{J.J.}},
\bauthor{\bsnm{Ogren}, \binits{P.V.}},
\bauthor{\bsnm{Zheng}, \binits{J.}},
\bauthor{\bsnm{Sohn}, \binits{S.}},
\bauthor{\bsnm{Kipper-Schuler}, \binits{K.C.}},
\bauthor{\bsnm{Chute}, \binits{C.G.}}:
\batitle{{Mayo clinical Text Analysis and Knowledge Extraction System (cTAKES): architecture, component evaluation and applications}}.
\bjtitle{Journal of the American Medical Informatics Association}
\bvolume{17}(\bissue{5}),
\bfpage{507}--\blpage{513}
(\byear{2010})
\end{barticle}
\endbibitem

\bibitem[\protect\citeauthoryear{Dligach et~al.}{2014}]{dligach2014discovering}
\begin{barticle}
\bauthor{\bsnm{Dligach}, \binits{D.}},
\bauthor{\bsnm{Bethard}, \binits{S.}},
\bauthor{\bsnm{Becker}, \binits{L.}},
\bauthor{\bsnm{Miller}, \binits{T.}},
\bauthor{\bsnm{Savova}, \binits{G.K.}}:
\batitle{Discovering body site and severity modifiers in clinical texts}.
\bjtitle{Journal of the American Medical Informatics Association}
\bvolume{21}(\bissue{3}),
\bfpage{448}--\blpage{454}
(\byear{2014})
\end{barticle}
\endbibitem

\bibitem[\protect\citeauthoryear{Xu et~al.}{2015}]{xu-etal-2015-uth}
\begin{bchapter}
\bauthor{\bsnm{Xu}, \binits{J.}},
\bauthor{\bsnm{Zhang}, \binits{Y.}},
\bauthor{\bsnm{Wang}, \binits{J.}},
\bauthor{\bsnm{Wu}, \binits{Y.}},
\bauthor{\bsnm{Jiang}, \binits{M.}},
\bauthor{\bsnm{Soysal}, \binits{E.}},
\bauthor{\bsnm{Xu}, \binits{H.}}:
\bctitle{{UTH}-{CCB}: The participation of the {S}em{E}val 2015 challenge {--} task 14}.
In: \bbtitle{Proceedings of the 9th International Workshop on Semantic Evaluation ({S}em{E}val 2015)},
pp. \bfpage{311}--\blpage{314}.
\bpublisher{Association for Computational Linguistics},
\blocation{Denver, Colorado}
(\byear{2015}).
\doiurl{10.18653/v1/S15-2052} .
\burl{https://aclanthology.org/S15-2052}
\end{bchapter}
\endbibitem

\bibitem[\protect\citeauthoryear{Xu et~al.}{2019}]{xu2019applying}
\begin{barticle}
\bauthor{\bsnm{Xu}, \binits{J.}},
\bauthor{\bsnm{Li}, \binits{Z.}},
\bauthor{\bsnm{Wei}, \binits{Q.}},
\bauthor{\bsnm{Wu}, \binits{Y.}},
\bauthor{\bsnm{Xiang}, \binits{Y.}},
\bauthor{\bsnm{Lee}, \binits{H.-J.}},
\bauthor{\bsnm{Zhang}, \binits{Y.}},
\bauthor{\bsnm{Wu}, \binits{S.}},
\bauthor{\bsnm{Xu}, \binits{H.}}:
\batitle{Applying a deep learning-based sequence labeling approach to detect attributes of medical concepts in clinical text}.
\bjtitle{BMC Medical Informatics and Decision Making}
\bvolume{19}(\bissue{5}),
\bfpage{1}--\blpage{8}
(\byear{2019})
\end{barticle}
\endbibitem

\bibitem[\protect\citeauthoryear{Shi et~al.}{2019}]{shi2019extracting}
\begin{barticle}
\bauthor{\bsnm{Shi}, \binits{X.}},
\bauthor{\bsnm{Yi}, \binits{Y.}},
\bauthor{\bsnm{Xiong}, \binits{Y.}},
\bauthor{\bsnm{Tang}, \binits{B.}},
\bauthor{\bsnm{Chen}, \binits{Q.}},
\bauthor{\bsnm{Wang}, \binits{X.}},
\bauthor{\bsnm{Ji}, \binits{Z.}},
\bauthor{\bsnm{Zhang}, \binits{Y.}},
\bauthor{\bsnm{Xu}, \binits{H.}}:
\batitle{Extracting entities with attributes in clinical text via joint deep learning}.
\bjtitle{Journal of the American Medical Informatics Association}
\bvolume{26}(\bissue{12}),
\bfpage{1584}--\blpage{1591}
(\byear{2019})
\end{barticle}
\endbibitem

\bibitem[\protect\citeauthoryear{Vaswani et~al.}{2017}]{Vaswani17}
\begin{bchapter}
\bauthor{\bsnm{Vaswani}, \binits{A.}},
\bauthor{\bsnm{Shazeer}, \binits{N.}},
\bauthor{\bsnm{Parmar}, \binits{N.}},
\bauthor{\bsnm{Uszkoreit}, \binits{J.}},
\bauthor{\bsnm{Jones}, \binits{L.}},
\bauthor{\bsnm{Gomez}, \binits{A.N.}},
\bauthor{\bsnm{Kaiser}, \binits{L.}},
\bauthor{\bsnm{Polosukhin}, \binits{I.}}:
\bctitle{Attention is all you need}.
In: \bbtitle{NIPS},
pp. \bfpage{5998}--\blpage{6008}
(\byear{2017})
\end{bchapter}
\endbibitem

\bibitem[\protect\citeauthoryear{Khandelwal and Britto}{2020}]{khandelwal-britto-2020-multitask}
\begin{bchapter}
\bauthor{\bsnm{Khandelwal}, \binits{A.}},
\bauthor{\bsnm{Britto}, \binits{B.K.}}:
\bctitle{Multitask learning of negation and speculation using transformers}.
In: \bbtitle{Proceedings of the 11th International Workshop on Health Text Mining and Information Analysis},
pp. \bfpage{79}--\blpage{87}.
\bpublisher{Association for Computational Linguistics},
\blocation{Online}
(\byear{2020}).
\doiurl{10.18653/v1/2020.louhi-1.9} .
\burl{https://aclanthology.org/2020.louhi-1.9}
\end{bchapter}
\endbibitem

\bibitem[\protect\citeauthoryear{Devlin et~al.}{2018}]{devlin2018bert}
\begin{botherref}
\oauthor{\bsnm{Devlin}, \binits{J.}},
\oauthor{\bsnm{Chang}, \binits{M.-W.}},
\oauthor{\bsnm{Lee}, \binits{K.}},
\oauthor{\bsnm{Toutanova}, \binits{K.}}:
Bert: Pre-training of deep bidirectional transformers for language understanding.
arXiv preprint arXiv:1810.04805
(2018)
\end{botherref}
\endbibitem

\bibitem[\protect\citeauthoryear{Yang et~al.}{2019}]{yang2019xlnet}
\begin{botherref}
\oauthor{\bsnm{Yang}, \binits{Z.}},
\oauthor{\bsnm{Dai}, \binits{Z.}},
\oauthor{\bsnm{Yang}, \binits{Y.}},
\oauthor{\bsnm{Carbonell}, \binits{J.}},
\oauthor{\bsnm{Salakhutdinov}, \binits{R.R.}},
\oauthor{\bsnm{Le}, \binits{Q.V.}}:
Xlnet: Generalized autoregressive pretraining for language understanding.
Advances in neural information processing systems
\textbf{32}
(2019)
\end{botherref}
\endbibitem

\bibitem[\protect\citeauthoryear{Liu et~al.}{2021}]{liu2021robustly}
\begin{bchapter}
\bauthor{\bsnm{Liu}, \binits{Z.}},
\bauthor{\bsnm{Lin}, \binits{W.}},
\bauthor{\bsnm{Shi}, \binits{Y.}},
\bauthor{\bsnm{Zhao}, \binits{J.}}:
\bctitle{A robustly optimized bert pre-training approach with post-training}.
In: \bbtitle{China National Conference on Chinese Computational Linguistics},
pp. \bfpage{471}--\blpage{484}
(\byear{2021}).
\bcomment{Springer}
\end{bchapter}
\endbibitem

\bibitem[\protect\citeauthoryear{Lee et~al.}{2020}]{lee2020biobert}
\begin{barticle}
\bauthor{\bsnm{Lee}, \binits{J.}},
\bauthor{\bsnm{Yoon}, \binits{W.}},
\bauthor{\bsnm{Kim}, \binits{S.}},
\bauthor{\bsnm{Kim}, \binits{D.}},
\bauthor{\bsnm{Kim}, \binits{S.}},
\bauthor{\bsnm{So}, \binits{C.H.}},
\bauthor{\bsnm{Kang}, \binits{J.}}:
\batitle{Biobert: a pre-trained biomedical language representation model for biomedical text mining}.
\bjtitle{Bioinformatics}
\bvolume{36}(\bissue{4}),
\bfpage{1234}--\blpage{1240}
(\byear{2020})
\end{barticle}
\endbibitem

\bibitem[\protect\citeauthoryear{Alsentzer et~al.}{2019}]{alsentzer2019publicly}
\begin{botherref}
\oauthor{\bsnm{Alsentzer}, \binits{E.}},
\oauthor{\bsnm{Murphy}, \binits{J.R.}},
\oauthor{\bsnm{Boag}, \binits{W.}},
\oauthor{\bsnm{Weng}, \binits{W.-H.}},
\oauthor{\bsnm{Jin}, \binits{D.}},
\oauthor{\bsnm{Naumann}, \binits{T.}},
\oauthor{\bsnm{McDermott}, \binits{M.}}:
Publicly available clinical bert embeddings.
arXiv preprint arXiv:1904.03323
(2019)
\end{botherref}
\endbibitem

\bibitem[\protect\citeauthoryear{Gu et~al.}{2021}]{gu2021domain}
\begin{barticle}
\bauthor{\bsnm{Gu}, \binits{Y.}},
\bauthor{\bsnm{Tinn}, \binits{R.}},
\bauthor{\bsnm{Cheng}, \binits{H.}},
\bauthor{\bsnm{Lucas}, \binits{M.}},
\bauthor{\bsnm{Usuyama}, \binits{N.}},
\bauthor{\bsnm{Liu}, \binits{X.}},
\bauthor{\bsnm{Naumann}, \binits{T.}},
\bauthor{\bsnm{Gao}, \binits{J.}},
\bauthor{\bsnm{Poon}, \binits{H.}}:
\batitle{Domain-specific language model pretraining for biomedical natural language processing}.
\bjtitle{ACM Transactions on Computing for Healthcare (HEALTH)}
\bvolume{3}(\bissue{1}),
\bfpage{1}--\blpage{23}
(\byear{2021})
\end{barticle}
\endbibitem

\bibitem[\protect\citeauthoryear{Lin et~al.}{2017}]{lin2017focal}
\begin{bchapter}
\bauthor{\bsnm{Lin}, \binits{T.-Y.}},
\bauthor{\bsnm{Goyal}, \binits{P.}},
\bauthor{\bsnm{Girshick}, \binits{R.}},
\bauthor{\bsnm{He}, \binits{K.}},
\bauthor{\bsnm{Doll{\'a}r}, \binits{P.}}:
\bctitle{Focal loss for dense object detection}.
In: \bbtitle{Proceedings of the IEEE International Conference on Computer Vision},
pp. \bfpage{2980}--\blpage{2988}
(\byear{2017})
\end{bchapter}
\endbibitem

\bibitem[\protect\citeauthoryear{Griffis et~al.}{2016}]{griffis2016quantitative}
\begin{barticle}
\bauthor{\bsnm{Griffis}, \binits{D.}},
\bauthor{\bsnm{Shivade}, \binits{C.}},
\bauthor{\bsnm{Fosler-Lussier}, \binits{E.}},
\bauthor{\bsnm{Lai}, \binits{A.M.}}:
\batitle{A quantitative and qualitative evaluation of sentence boundary detection for the clinical domain}.
\bjtitle{AMIA Summits on Translational Science Proceedings}
\bvolume{2016},
\bfpage{88}
(\byear{2016})
\end{barticle}
\endbibitem

\bibitem[\protect\citeauthoryear{Pearson}{1900}]{pearson1900x}
\begin{barticle}
\bauthor{\bsnm{Pearson}, \binits{K.}}:
\batitle{X. on the criterion that a given system of deviations from the probable in the case of a correlated system of variables is such that it can be reasonably supposed to have arisen from random sampling}.
\bjtitle{The London, Edinburgh, and Dublin Philosophical Magazine and Journal of Science}
\bvolume{50}(\bissue{302}),
\bfpage{157}--\blpage{175}
(\byear{1900})
\end{barticle}
\endbibitem

\bibitem[\protect\citeauthoryear{Webson and Pavlick}{2022}]{webson-pavlick-2022-prompt}
\begin{bchapter}
\bauthor{\bsnm{Webson}, \binits{A.}},
\bauthor{\bsnm{Pavlick}, \binits{E.}}:
\bctitle{Do prompt-based models really understand the meaning of their prompts?}
In: \bbtitle{Proceedings of the 2022 Conference of the North American Chapter of the Association for Computational Linguistics: Human Language Technologies},
pp. \bfpage{2300}--\blpage{2344}.
\bpublisher{Association for Computational Linguistics},
\blocation{Dublin, Ireland}
(\byear{2022}).
\burl{https://aclanthology.org/2022.naacl-main.167.pdf}
\end{bchapter}
\endbibitem

\bibitem[\protect\citeauthoryear{Lehman et~al.}{2023}]{lehman2023we}
\begin{botherref}
\oauthor{\bsnm{Lehman}, \binits{E.}},
\oauthor{\bsnm{Hernandez}, \binits{E.}},
\oauthor{\bsnm{Mahajan}, \binits{D.}},
\oauthor{\bsnm{Wulff}, \binits{J.}},
\oauthor{\bsnm{Smith}, \binits{M.J.}},
\oauthor{\bsnm{Ziegler}, \binits{Z.}},
\oauthor{\bsnm{Nadler}, \binits{D.}},
\oauthor{\bsnm{Szolovits}, \binits{P.}},
\oauthor{\bsnm{Johnson}, \binits{A.}},
\oauthor{\bsnm{Alsentzer}, \binits{E.}}:
Do we still need clinical language models?
arXiv preprint arXiv:2302.08091
(2023)
\end{botherref}
\endbibitem

\end{thebibliography}

\end{document}